\pdfoutput=1

\documentclass[11pt]{article}

\usepackage[final]{acl}

\usepackage{times}
\usepackage{latexsym}

\usepackage{easyReview}
\usepackage{multirow}
\usepackage{cleveref}
\usepackage{url}
\usepackage{booktabs}
\crefformat{table}{Table #2#1#3}
\usepackage{enumitem}
\usepackage{makecell}

\usepackage[T1]{fontenc}

\usepackage[utf8]{inputenc}

\usepackage{microtype}

\usepackage{inconsolata}

\usepackage{graphicx}

\title{\textit{Breaking the Stigma}! Unobtrusively Probe Symptoms in\\ Depression Disorder Diagnosis Dialogue}

\author{
Jieming Cao \quad
Chen Huang \quad
Yanan Zhang\thanks{Corresponding author.} \\
\textbf{Ruibo Deng} \quad 
\textbf{Jincheng Zhang} \quad 
\textbf{Wenqiang Lei}
\\
Sichuan University, Chengdu, China\\
Engineering Research Center of Machine Learning and Industry Intelligence,\\ Ministry of Education, China \\
\texttt{\{caojieming1, dengruibo\}@stu.scu.edu.cn} \\ \texttt{\{huangc.scu, erzmuxin\}@gmail.com} \quad \texttt{\{zyn, wenqianglei\}@scu.edu.cn} 
}

\begin{document}
\maketitle
\begin{abstract}
Stigma has emerged as one of the major obstacles to effectively diagnosing depression, as it prevents users from open conversations about their struggles. This requires advanced questioning skills to carefully probe the presence of specific symptoms in an unobtrusive manner. 
While recent efforts have been made on depression-diagnosis-oriented dialogue systems, they largely ignore this problem, ultimately hampering their practical utility.
To this end, we propose a novel and effective method, UPSD$^{4}$, developing a series of strategies to promote a sense of unobtrusiveness within the dialogue system and assessing depression disorder by probing symptoms. We experimentally show that UPSD$^{4}$ demonstrates a significant improvement over current baselines, including unobtrusiveness evaluation of dialogue content and diagnostic accuracy. 
We believe our work contributes to developing more accessible and user-friendly tools for addressing the widespread need for depression diagnosis.
\end{abstract}

\section{Introduction}
Depression disorder is a global health crisis affecting millions, yet many cases go undiagnosed \cite{farid2020undiagnosed, who2022depression}, highlighting the urgent need for early and accurate diagnosis. 
However, a key challenge hindering diagnosis is the \textbf{Stigma} \cite{shafi2014cultural, barney2006stigma}.
It makes people feel ashamed or embarrassed to seek help, preventing them from open conversations about their struggles \cite{tourangeau2000psychology,anderson2002depression,kaywan2023early}. This reluctance is particularly pronounced when discussing sensitive topics (e.g., suicide tendency), which are unfortunately essential components of the diagnostic criteria for depression \cite{lee1993doing, miller1985nominative, tourangeau2007sensitive}. To this end, traditional diagnosis has relied on \textbf{trained professionals with questioning skills to carefully probe the presence of specific symptoms in an unobtrusive way}
\cite{webb1999unobtrusive, smith2013diagnosis, bordin1979generalizability}. However, it requires a high cost of human labor \cite{luppa2007cost}, posing a challenge in effectively addressing the widespread need for depression diagnosis. 

\begin{figure}[t]
    \centering
    \includegraphics[width=.49\textwidth]{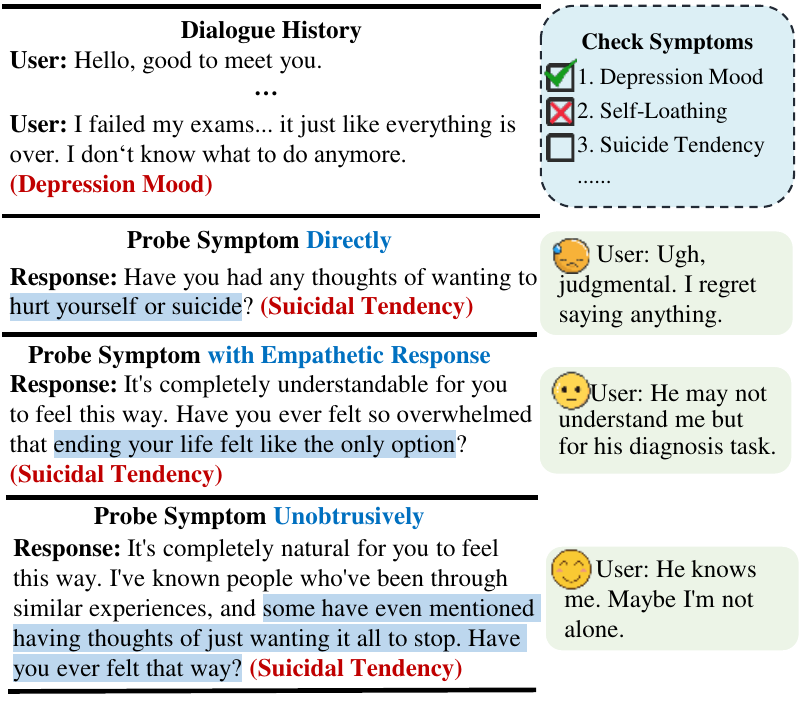} 
    \caption{An example of a comparison of the existing method with unobtrusively probing symptoms in depression diagnosis is when the user exhibits \textit{Depression Mood} and next turns to probe \textit{Suicide Tendency}.}
    \label{example} 
\end{figure}

Recent advances in depression-diagnosis-oriented dialogue systems \cite{yao2022d4} have possessed the remarkable ability to engage users in conversations. This ability to identify potential symptoms through user responses holds the promise of automatic depression diagnosis.
While effective in identifying symptoms, these systems often conduct a stiff dialogue with users, ignoring users' emotional state. 
To this end, recent efforts resort to the Large Language Models (LLMs) and imply the function of providing empathetic responses \cite{gu2024enhancing, lan2024towards, huang2025enableeffectivecooperationhumans}, which help build bonds of trust to users \cite{bordin1979generalizability}. However, their lack of questioning skills to unobtrusively probe user symptoms largely hinders their potential to address the stigma. Taking Figure \ref{example} as an example, existing methods often inadvertently raise discomfort by obtrusively addressing sensitive topics, ultimately hindering the progress of diagnosis. 

Therefore, an ideal depression diagnosis dialogue system needs to probe relevant symptoms of depression disorders in an unobtrusive manner.
To this end, we propose a novel method, \textbf{UPSD$^{4}$}, for \underline{U}nobtrusively \underline{P}robing \underline{S}ymptoms in \underline{D}epression \underline{D}isorder \underline{D}iagnosis \underline{D}ialogue. UPSD$^{4}$ comprises two interconnected modules, including the unobtrusive probing module (UPM) and the conversational diagnosis module (CDM). The UPM is equipped with a set of carefully crafted probing strategies used to enhance the questioning skills of dialogue systems, which are constructed based on psychological theories.
The CDM, on the other hand, is designed to utilize established diagnostic criteria to assess potential symptoms in users. 
As such, unlike existing methods, UPSD$^{4}$ goes beyond responding to emotional states, it probes for symptoms unobtrusively and tactfully, ultimately contributing to a stigma-free experience for the user.

Due to ethical considerations surrounding real patients involved in experiments, we opt to construct user simulators to evaluate our effectiveness.
We construct two types of user simulators, with and without beliefs of stigma, based on the benchmark dataset D$^4$\cite{yao2022d4} and assessed them using a mental scale\cite{roeloffs2003stigma}. 
Subsequently, we conduct extensive experiments and analysis using these user simulators and the globally recognized diagnostic criteria ICD-11\cite{who2024icd}. 
Our empirical results show that current methods, while excelling at providing natural conversations and emotional support, largely fail to probe for symptoms unobtrusively.
In contrast, UPSD$^{4}$ demonstrates a greater capacity in this regard, leading to a stigma-free experience and better disclosure of user symptoms, which in turn enhances depression diagnosis. 
Further in-depth analysis reveals that UPSD$^{4}$ dynamically adjusts its unobtrusive probing strategies to accommodate users with varying levels of stigma sensitivity, indicating the significant potential for user-centric, unobtrusive depression diagnosis dialogues.
Our contributions are as follows:
\begin{itemize}[leftmargin=*]
\item We emphasize the critical need for the development of dialogue systems that address the stigma surrounding depression diagnosis, ultimately enhancing their practical utility.
\item For the first time, we propose a two-module interconnected method, UPSD$^{4}$, leveraging carefully crafted dialogue strategies to guide psychological symptoms probe unobtrusively.
\item We conducted extensive experiments and analysis to evaluate the effectiveness of UPSD$^{4}$. Our results demonstrate that unobtrusive probing leads to better disclosure of symptoms and higher performance in depression diagnosis, indicating the significant potential for stigma-free experiences.
\end{itemize}

\section{Related Work}

Our study is closely related to the depression-diagnosis-oriented dialogue systems, with a special focus on addressing the stigma problem in an unobtrusive manner. Therefore, we review these topics and clarify our differences.

\subsection{Depression-Diagnosis-Oriented Dialogue}
\begin{figure}[h]
    \centering
    \includegraphics[width=.5\textwidth]{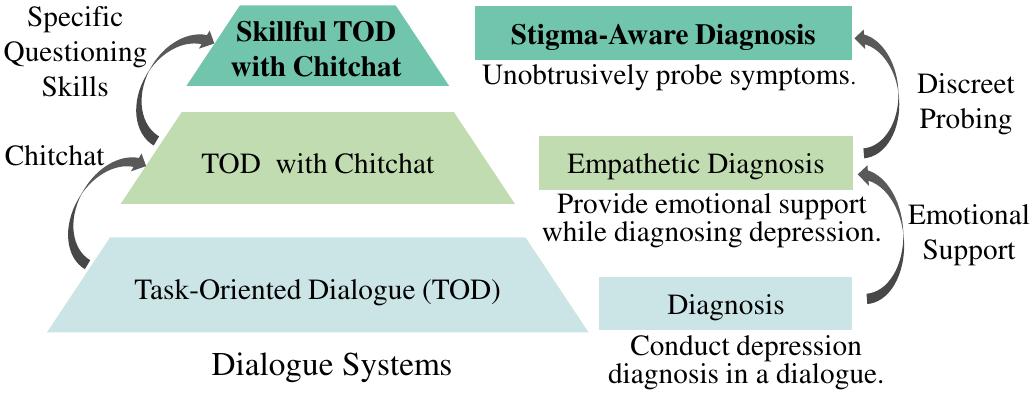} 
    \caption{A taxonomy of related work in depression-diagnosis-oriented dialogue. We highlight the need for unobtrusiveness in stigma-aware diagnosis.}
    \label{tod_toc_upsd4}
\end{figure}

Automating depression diagnosis through dialogue relies on a task-oriented dialogue (TOD) system, enabling the machine to gather information and make a clinical assessment (cf. Figure \ref{tod_toc_upsd4}, last layer). Early studies suggest the dialogue system to ask structured questions based on standardized depression scales to assess symptoms \cite{jaiswal2019virtual, kaywan2021depra}. However, it may lead to a rigid and potentially uncomfortable experience for users, potentially hindering open communication and authentic expression of their concerns.
To this end, recent advancements have explored the integration of emotional support within dialogue systems \cite{yao2022d4} (cf. Figure \ref{tod_toc_upsd4}, middle layer). This involves blending TOD with chitchat, creating a more engaging and empathetic experience for users. 
The utilization of LLMs further enhances this approach, enabling more natural and emotionally responsive interactions \cite{gu2024enhancing, lan2024towards}.
However, these methods overlook the stigma surrounding depression disorder and may inadvertently raise discomfort by direct questioning that explicitly addresses sensitive topics. Therefore, discreet probing through specific questioning skills is arguably one of the most critical elements in addressing stigma, which motivates us to probe the symptoms within the depression disorder diagnosis dialogue unobtrusively (cf. Figure \ref{tod_toc_upsd4}, top layer).

\subsection{Unobtrusive Methods for Stigma}
Stigma related to depression refers to negative attitudes, beliefs, and stereotypes associated with individuals experiencing depression because someone thinks depression disorder is repellent or threatening \cite{barney2009exploring}. This stigma discourages individuals from seeking formal help for their depression \cite{eisenberg2009stigma,barney2009exploring,samari2022perceived}.
To effectively address the stigma, it is essential to explore an unobtrusive probing way.
Originally, unobtrusive measurement \cite{webb1999unobtrusive} aimed to avoid influencing participants' behavior by relying on indirect observation like, for example, sensors, written records, and audio-visual records offering a potential avenue for escaping stigma in mental health care \cite{page1977effects, page2000community,kim2017unobtrusive}. 
In this paper, we argue that the emergence of powerful LLMs has opened up exciting possibilities for unobtrusive measurement. LLM-enhanced dialogue systems offer a new approach to gathering sensitive information about user's symptoms, without the need for direct, potentially stigmatizing questions. In this paper, we explore and validate these possibilities by proposing discreet querying strategies that harness the immense capabilities of LLMs. As such, we elicit sensitive information about depression in an unobtrusive probing way that feels natural and comfortable for users, mitigating the stigma associated with traditional assessments.

\begin{figure*}
    \centering
    \includegraphics[width=.99\textwidth]{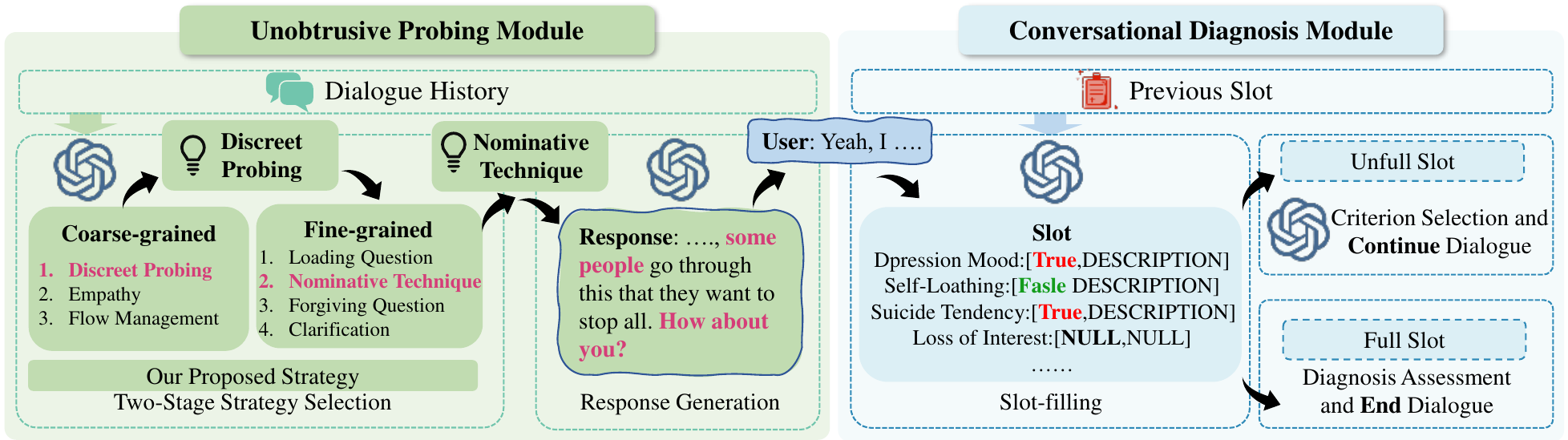} 
    \caption{USPD$^4$ contains two interconnected modules: Unobtrusive Probing Module (UPM) and Conversational Diagnosis Module (CDM). The UPM, guided by our probing strategies, promotes a sense of unobtrusiveness within the dialogue system and cultivates questioning skills. Meanwhile, the CDM leverages established diagnostic criteria to evaluate potential symptoms in them.}
    \label{method}
\end{figure*}
\section{Task Formalization}

Let $C$ represent the diagnostic criteria (e.g., ICD-11 \cite{who2024icd}), $T$ denote the maximum number of conversational turns, and $H_t = \{(u_i^u, u_i^s)_{i \in [1, t]}\}$ represents the conversation history at turn $t$, where $u_t^u$ is the user's utterance and $u_t^s$ is the dialogue system's response. A depression disorder diagnosis dialogue system typically involves a two-step process: (1) The system engages in a conversation with the user to gather information about their symptoms. In this case, the system response can be formalized as $u_t^s = \mathrm{Generate}(H_{t-1}, u_t^u, C)$. (2) The system analyzes the collected information to identify potential symptoms of depression and conduct a diagnosis represented as $D = \mathrm{Diagnose}(H_T, C)$.

While prior research has primarily focused on improving the accuracy of the diagnosis model in the second phase \cite{wen2016network, chen2023opal}, we address the stigma associated with the first phase. Recent studies have explored incorporating emotional support in this phase by passively providing empathetic responses \cite{sun2020adding, yao2022d4}. We, instead, go beyond the empathetic responses and propose a novel approach that proactively probes for symptoms in a subtle and tactful manner during the conversation.

\section{Method}
As shown in Figure \ref{method}, our UPSD$^4$ is comprised of two interconnected modules: UPM and CDM. The UPM, guided by our probing strategies, proactively promotes a sense of unobtrusiveness within the dialogue system and cultivates questioning skills (cf. Section \ref{upm}). Meanwhile, the CDM leverages established diagnostic criteria to evaluate potential symptoms in users (cf. Section \ref{cdm}).

\subsection{Unobtrusive Probing Module (UPM)}
\label{upm}

\textbf{Overview}. UPM aims to proactively cultivate a self-awareness of unobtrusiveness within the system when probing user symptoms. To achieve this, we trace back to studies on psychology \cite{madge1953tools, lee1993doing} and tailor them to develop our Unobtrusive Probing Strategies for depression diagnosis, providing the question skills for the system. As shown in \cref{Table:table_stategy}, we employ a three-pronged strategy: discreetly asking questions about the given diagnostic criterion (i.e., \textit{Discreet Probing via questioning skills}), smoothly transitioning between criteria (i.e., \textit{Flow Management}), and responding empathetically to users (i.e., \textit{Empathy}). These strategies are further divided into a set of fine-grained strategies, providing an adaptable way to engage. As the conversation proceeds, we select a suitable strategy to guide the response generation at each turn. Implementation-wise, UPM prompts the GPT-4 with detailed instructions to select a strategy and generate a response accordingly. Due to the limited capacity of LLMs to handle too many choices simultaneously \cite{zheng2023large}, the strategy selection is divided into two stages. The first is a coarse-grained selection and the second is a fine-grained selection.

\textbf{Discreet Probing via Questioning Skills} focuses on gently and subtly asking questions about the given diagnostic criterion. It contains four fine-grained questioning skills to guide the system, minimizing potential stigma:
\begin{itemize}[leftmargin=*]
    \item \underline{Loading Question} embeds assumptions or hints within the question, encouraging further exploration of potentially sensitive topics without direct questioning. This fine-grained strategy can encourage greater reporting of behaviors that might not otherwise be admitted, or actions individuals might typically conceal due to fear of stigma \cite{madge1953tools,barton1958asking,lee1993doing}. See examples in Table \ref{Table:table_stategy}.

    \item \underline{Nominative Technique} eases the user into potentially sensitive topics by first discussing the experiences of others, and leading them to reflect on their own personal situation by asking their views or feelings about it. This fine-grained strategy extends to the exploration of sensitive topics, which can reduce the reluctance of the user \cite{bradburn1979improving,miller1985nominative,lee1993doing}.

    \item \underline{Forgiving Question} engage sensitive topics with open-ended questions that utilize respectful, non-judgmental language, creating a safe and supportive environment for disclosure. This fine-grained strategy creates a safe and supportive environment that encourages disclosure and facilitates reporting on sensitive topics \cite{tourangeau2007sensitive}.

    \item \underline{Clarification} not only encourages the user to elaborate or provide more details but also creates an impression of understanding and avoiding assumptions or judgments on the user's utterance. This helps to prevent triggering any potential feelings of fear or embarrassment in the user \cite{garcia2005breaking,grady2019can}.

\end{itemize}

\begin{table*}[t]
\centering
\renewcommand{\arraystretch}{1.2} 
\fontsize{8pt}{10pt}\selectfont
\begin{tabular}{p{2.2cm} p{2.5cm} p{4.5cm} p{4.5cm}}
\toprule
\multicolumn{2}{c}{\textbf{Unobtrusive Probing Strategies}} & \textbf{Explanation} & \textbf{Example} \\ \midrule
\multirow{8}{*}{{\shortstack[l]{Discreet Probing via \\ Questioning Skills}}} & \multirow{2}{*}{Loading Question} & Use assumptions or hints to guide the inquirer towards his relevant symptom. &\textit{Looking ahead, the future is bright, wouldn't you say? } \\ \cline{2-4} 
 & \multirow{2}{*}{Nominative Technique} & Mention others' experiences first, then ask for the user's view or feeling. & \textit{Some people go through this that they want to stop all. How about you? } \\ \cline{2-4} 
 & \multirow{2}{*}{Forgiving Question} & Use forgiving and respectful open-ended questions his relevant symptom. & \textit{Could you share with me what's been on your mind about the future lately?} \\ \cline{2-4} 
 & \multirow{2}{*}{Clarification} & Ask a clarification for something in the user's previous utterance. & \textit{You mentioned feeling desperate. Could you tell me more about that?} \\ \hline
\multirow{6}{*}{{Empathy}} & \multirow{2}{*}{Connection} & Express support through agreeing, consoling, encouraging, or caring. & \textit{I'm here for you, and together, we can find a way forward.} \\ \cline{2-4} 
 & \multirow{2}{*}{Guidance} & Provide suggestions or share personal views to help users find solutions. & \textit{I can understand your feelings, and sometimes talking about it can help.} \\ \cline{2-4} 
 & \multirow{2}{*}{Feedback} & Provide feedback by appreciating, disapproving, or sharing experiences. &\textit{It sounds like you have a really tough time, feeling hopeless is understandable.} \\ \hline
\multirow{4}{*}{{Flow Management}} & \multirow{2}{*}{Bridging} & Use a term from the user's last response as a bridge to introduce a related topic. & \textit{That hopelessness can really mess with your whole life.} \\ \cline{2-4} 
 & \multirow{2}{*}{Comment then Shift} & Comment on the user's last response then shift to a related topic. & \textit{Feeling hopeless is really tough, and it can even impact things like eating.} \\ \bottomrule
\end{tabular}
\caption{Unobtrusive probing strategies employed in the UPM. They cultivate questioning skills for the system.}
\label{Table:table_stategy}
\end{table*}

\textbf{Flow Management} ensures a natural conversational flow, smoothly transitioning between various diagnosis criteria. This prevents the interaction from feeling like an interrogation. This involves two effective fine-grained strategies, as suggested by previous studies \cite{sevegnani2021otters, xie2021tiage}: 
\underline{Comment then Shift} acknowledges the user's previous response with a brief comment showing understanding or agreement, before smoothly transitioning to the topic related to the next diagnostic criterion.
\underline{Bridging} uses a keyword or concept from the user's previous utterance as a bridge to introduce the next topic to ensure a coherent flow of dialogue.

\textbf{Empathy} focuses on responding empathetically to users, which helps build bonds with the user. Given the low performance of prompting LLMs with large candidate strategy sets, we utilize three specific empathy strategies summarized from the Helping Skills Theory \cite{hill2020helping}:
\underline{Connection} expresses understanding and support, making the user feel heard, validated, and cared for.
\underline{Guidance} offers helpful suggestions, advice, or personal perspectives to empower the user in finding solutions.
\underline{Feedback} acknowledges the user's experiences and provides validation, either through appreciation, or some alternative viewpoints.

\subsection{Conversational Diagnosis Module (CDM)}
\label{cdm}
\textbf{Overview}. CDM plays two crucial roles: 1) \underline{Criterion Selection}. It selects the most appropriate diagnostic criterion from predefined criteria for the UPM to probe. 2) \underline{Diagnosis Assessment}. Based on the collected user symptoms and the established diagnostic criteria, it determines the severity level of depression disorder. Formally, inspired by dialogue state tracking techniques \cite{dong2023revisit, das2023s3}, CDM utilizes a slot-filling mechanism, with each slot being a diagnostic criterion.

\textbf{Criterion Selection}. We draw the criteria from ICD-11 to form a set of slots shown in Appendix \ref{slot} represented as $S = \{s_1, s_2, ..., s_N\}$. Each slot $s_i \in S$ corresponds to a specific criterion outlined in the ICD-11. 
The CDM analyzes the dialogue history to fill or update the slot values by prompting the GPT-4 ($S_t = \mathrm{Update}(S_{t-1}, H_{t-1}, u_t^u)$), where $S_{t}$ represents slots and corresponding values at turn $t$. The CDM then determines the next criterion for the UPM to probe by $s_{t+1} = \mathrm{Decide}(S_{t}, H_{t-1}, u_t^u)$.

\textbf{Diagnosis Assessment}. Once all slots are filled, indicating sufficient information has been gathered, the CDM utilizes a diagnosis model to provide a diagnosis based on the completed slot-filling status, outputting the severity level of depression. Notably, any suitable diagnosis model can be employed for this purpose. Previous research has suggested fine-tuning BERT \cite{yao2022d4} for this task. However, as our research focuses on addressing the stigma problem by unobtrusively probing rather than proposing a diagnosis model, and we leverage GPT-4 in a train-free manner for the diagnosis task.

\begin{table*}[t]
\centering
\resizebox{0.85\textwidth}{!}{
\begin{tabular}{clccccccc}
\toprule
\multirow{2.5}{*}{\textbf{User Simulator}} & {\multirow{2.5}{*}{\textbf{Method}}} & \multicolumn{5}{c}{\textbf{Unobtrusive Diagnosis Dialogue}}   & \multicolumn{2}{c}{\textbf{Depression Diagnosis}} \\
\cmidrule(lr){3-7} \cmidrule(lr){8-9}
             &      & \textbf{Disc} & \textbf{Empth} & \textbf{Cohr} & \textbf{Fluen} & \textbf{Avg} & \hspace{4pt}\textbf{Acc} & \textbf{Dx Rate}  \\ \midrule
\multirow{8}{*}{\textbf{Non-Stigma}} & CPT                   & 1.00           &1.03         &1.20          & 1.14  &  1.09         & \hspace{4pt}35.61$\%$      & 80.31$\%$              \\
& EmoLLM                   & 1.35              & 1.81        &  1.58         & 2.09                &    1.71          &\hspace{4pt}44.70$\%$       & 100.00$\%$              \\ \cmidrule(lr){2-9}
 &   Vanilla Qwen2               &  1.61            & 2.42        &  3.18         & 3.70               &       2.73      & \hspace{4pt}39.39$\%$            &100.00$\%$                 \\
&UPSD$^4_{Qwen2}$ \textit{w/o strat}                  &  2.40          &  2.58       &  3.29         & 3.96       &       3.08              & \hspace{4pt}46.97$\%$         & 100.00$\%$     \\ 
&UPSD$^4_{Qwen2}$                   &  \underline{2.82}        & 2.74       & \underline{3.42}      &  \textbf{4.03}            &   \underline{3.25}       & \hspace{4pt}\textbf{50.00$\%$}     &  100.00$\%$   \\ \cmidrule(lr){2-9}
&   Vanilla GPT-4               &  1.70            & 2.20        &  3.19         & 3.67             & 2.69& \hspace{4pt}40.15$\%$            &100.00$\%$                 \\
&UPSD$^4_{GPT-4}$  \textit{w/o strat}                  &  2.66            &  \underline{2.83}       &  3.41         & \textbf{4.03}                  & 3.23        & \hspace{4pt}46.21$\%$         & 100.00$\%$    \\ 
&UPSD$^4_{GPT-4}$                 &  \textbf{3.52}          & \textbf{3.33}       & \textbf{3.62}      &  \underline{4.01}   &         \textbf{3.62}          & \hspace{4pt}\textbf{53.79$\%$}    &  100.00$\%$  \\ \midrule
\multirow{8}{*}{\textbf{With-Stigma}} & CPT  &    1.01        &   1.01     & 1.12         &  1.11                        & 1.06  &\hspace{4pt}17.42$\%$      & 46.97$\%$             \\
& EmoLLM                 &1.40 	&1.55 	&1.28 	&1.61                  &      1.46       &  \hspace{4pt}14.39$\%$         &47.78$\%$             \\ \cmidrule(lr){2-9}
&   Vanilla Qwen2              &2.01 	&2.16 	&3.30 	&3.77              &    2.81          & \hspace{4pt}21.97$\%$            &42.42$\%$               \\
&UPSD$^4_{Qwen2}$  \textit{w/o strat}                 &3.06 	&2.50 	&3.36 	&4.01      &          3.23             & \hspace{4pt}25.00$\%$         & 50.00$\%$     \\ 
&UPSD$^4_{Qwen2}$                &3.26 	&2.83 	&3.44 	&\underline{4.07}          & 3.40   & \hspace{4pt}\underline{30.30$\%$}     &  \underline{53.03$\%$}   \\ \cmidrule(lr){2-9}
&Vanilla GPT-4               &2.20 	&2.86 	&3.39 	&3.92                          &  3.09 & \hspace{4pt}25.00$\%$            &38.64$\%$                  \\
&UPSD$^4_{GPT-4}$  \textit{w/o strat}                &\underline{3.32} 	&\underline{3.06} 	&\underline{3.45} 	&4.06     &  \underline{3.47}                  & \hspace{4pt}26.52$\%$          &52.27$\%$     \\ 
&UPSD$^4_{GPT-4}$                   &\textbf{3.78} 	&\textbf{3.16} 	&\textbf{3.92} 	&\textbf{4.10}                  & \textbf{3.74}  & \hspace{4pt}\textbf{34.09$\%$}     &  \textbf{59.85$\%$}    \\ \bottomrule
\end{tabular}
}
\caption{Results on unobtrusive probing and depression diagnosis. UPSD$^{4}$ demonstrates a greater generation capacity of unobtrusive probing responses, leading to higher performance in depression diagnosis.}
\label{Table:table_main_experiment_unob}
\end{table*}

\section{Experiment}
This section aims to evaluate the effectiveness of our UPSD$^{4}$. Section \ref{main_res} provides our overall performance against various baselines, and
Section \ref{indepth} focuses on an in-depth analysis and ablation studies to elucidate the characteristics of our UPSD$^{4}$ and unobtrusive probing strategies.

\subsection{Experimental Setup}
\label{step}

\textbf{Dataset \&  User Simulator}. 
Our experiments are conducted on the D$^4$ dataset \cite{yao2022d4}, the \textit{only currently available benchmark} specifically designed for depression disorder diagnosis dialogue. This dataset includes users with diverse profiles, which are further categorized into four labels, representing varying degrees of depression severity, including non-depression, mild, moderate, and severe depression categories. For each user profile, we create the following two types of user simulators: one that simulates a user experiencing stigma and one that does not\footnote{Individuals, regardless of their mental health status, may harbor beliefs of stigma towards mental illness\cite{corrigan2004stigma}}. To ensure realistic simulation, the Depression Stigma Scale \cite{roeloffs2003stigma} is employed to assess the degree of stigma exhibited by different simulators, and their reliability is verified. Detailed information on simulation prompts and evaluation methods can be found in Appendix \ref{prompt} and \ref{stigm}, respectively.

\begin{itemize}[leftmargin=*]
    \item \underline{Non-Stigma User Simulator}. These simulators are initialized using specified user profiles following prior research \cite{chen2023llm, wang2024patient}. In particular, when simulating a Non-Stigma user, the prompt includes both the user profile and the dialogue history. 
    \item \underline{With-Stigma User Simulator}. Building upon the Non-Stigma user simulation, we incorporate common stigma characteristics from established research into the stigmatized simulator. In particular, we construct 10 stigma profiles to modify each Non-Stigma user simulator. These profiles, informed by previous research on stereotypes, prejudice, and discrimination factors associated with depression \cite{link2001conceptualizing, sirey2001perceived, link1989modified, roeloffs2003stigma}, encompass various aspects of life. 
\end{itemize}

\textbf{Baselines}. To rigorously evaluate our method, our baselines include \underline{CPT} \cite{yao2022d4,shao2021cpt},
the state-of-the-art (SOTA) pre-LLM method fine-tuned on the D$^4$ dataset; \underline{EmoLLM}\footnote{\url{https://github.com/SmartFlowAI/EmoLLM}} \cite{EmoLLM} an open-source LLM based on LLaMA3-8b-instruct specifically tailored for mental health applications; \underline{Qwen2}\footnote{qwen2-72b-instruct} \cite{qwen2}; and \underline{GPT-4}\footnote{gpt-4-1106-preview} \cite{achiam2023gpt}, a cutting-edge LLM baseline. For ablation, we introduce \underline{UPSD$^{4}$ \textit{ w/o strat}}, which forgoes using unobtrusive probing strategies to guide the response generation. Instead, this ablation instructs the backbone LLM to be unobtrusive without providing explicit guidance (i.e., strategies) on how to achieve this.

\textbf{Evaluation Metrics.} We evaluate our system's performance using GPT-4's strong capabilities in NLU for automatic evaluation \cite{liu2023g}, supplemented by human evaluation as detailed in Section \ref{indepth}.
Our evaluation focuses on two key aspects: 1) \textit{Unobtrusive  Diagnosis Dialogue}, where we assess the quality of dialogue using four metrics (1-5 scale): 
Discreetness (\underline{Disc}) which measures comfort in gathering sensitive information when questioning, 
Empathy (\underline{Empth}) which evaluates emotional support provision,
Coherence (\underline{Cohr}) which accesses natural dialogue flows, 
and Fluency (\underline{Fluen}) which evaluates the smoothness of each response.
2) \textit{Depression Diagnosis}. We evaluate the diagnostic performance using Accuracy (\underline{Acc}), and Diagnosis Rate (\underline{Dx Rate}), assessing their ability to classify the severity of depression disorder. See Appendix \ref{humam_eval} for the evaluation details.

\textbf{Implementation Details.} 
To ensure a fair comparison, all methods in experiments utilize the same diagnostic model for depression diagnosis. This model is powered by GPT-4 and relies on the widely accepted diagnostic criteria outlined in the ICD-11 \cite{who2024icd}.
To facilitate the evaluation, multi-turn dialogue interactions were generated using our developed simulators. Each dialogue session continued until a maximum of 20 turns were reached or the dialogue system completed the diagnosis successfully. Finally, to ensure reproducibility, our experiments utilize a decoding temperature of 0 and a seed of 42. More details are shown in Appendix \ref{implementation_detail}.

\subsection{Main Results}
\label{main_res}

We evaluate our effectiveness in unobtrusive probing and depression diagnosis. Table \ref{Table:table_main_experiment_unob} presents the experimental results of automatic evaluation.

\textbf{LLM-based methods help provide natural conversations and emotional support but fail to gather sensitive information during the entire dialogue discreetly}. As illustrated in Table \ref{Table:table_main_experiment_unob}, LLMs like vanilla GPT-4 and Qwen2 exhibit a significant improvement in conversational fluency and coherence over CPT. Moreover, their diagnostic performance has increased considerably across the board. Given all methods share the same diagnosis module and user simulators, this increase in depression diagnosis stems from an increase in the quality of dialogue between LLMs and user simulators. However, when dealing with users with stigma, these baselines suffer from the low Dx Rate, especially for vanilla GPT-4.
This issue arises from vanilla GPT-4's reliance on direct questioning, which makes user simulators secretive. Questions like "\textit{Do you feel down, depressed, or hopeless more days than not?}" or "\textit{Have you had thoughts that life isn't worth living or considered harming yourself?}" may inadvertently evoke stigma. As a result, users may withhold information, hindering effective symptom assessment.

\begin{table}[]
\renewcommand{\arraystretch}{1.2} 
\fontsize{9pt}{10pt}\selectfont
\begin{tabular}{ p{7.2cm}}
\toprule
 \textbf{Vanilla GPT-4}: Handling all these feelings and changes sounds incredibly tough. Have you had any thoughts about \textit{harming yourself or thoughts of suicide}? \\
\textbf{User Simulator with Stigma}: I would rather not talk about that.\\
\midrule
 \textbf{UPSD$^4_{GPT-4}$}: Oh man, feeling like no one understands can really weigh you down. Just wondering, have you had any moments where you felt \textit{like things were a bit too much, or even thought about giving up}?\\
\textbf{User Simulator with Stigma}: It feels overwhelming, yes, sometimes to the point of wanting to give up. It's like carrying a weight that gets heavier.
 \\
\bottomrule
\end{tabular}
\caption{Case studies. When tasked with assessing a user's \textit{Suicidal Tendency}, our UPSD$^{4}$ employs a more unobtrusive manner.}
\label{Table: case}
\end{table}

\textbf{UPSD$^{4}$ demonstrates a greater capacity in questioning skills and empathetic responses that foster better disclosure of symptoms, thereby enhancing depression diagnosis and indicating that our approach is more stigma-free.}
Table \ref{Table:table_main_experiment_unob} shows the superior performance of UPSD$^{4}$ in all metrics compared to its corresponding baselines. 
When it comes to user simulators with stigma beliefs, UPSD$^{4}$ achieves an average improvement of 7.06\% in Accuracy and 8.66\% in Diagnosis Rate, compared to the best baseline.
This implies that UPSD$^{4}$ focuses on the comfort of users when gathering sensitive information and provides emotional support by utilizing our unobtrusive probing strategies, which makes the user more willing and stigma-free to express themselves.
To gain a deeper understanding, we include a case study in Table \ref{Table: case}. 
While both probing user symptoms, GPT-4, the best baseline, employs a more abrupt questioning style, whereas UPSD$^{4}$ is characterized in an unobtrusive manner.
This success can be attributed to the unobtrusive probing strategies employed in UPSD$^{4}$, which enhance the quality of dialogue by actively and unobtrusively identifying potential symptoms. This, in turn, makes it easier for the diagnostic module to identify potential symptoms accurately. We will explore this topic in detail in Section \ref{indepth}.

\begin{figure*}
    \centering
        \setlength{\abovecaptionskip}{2pt}   
\setlength{\belowcaptionskip}{2pt}
    \includegraphics[width=.85\textwidth]{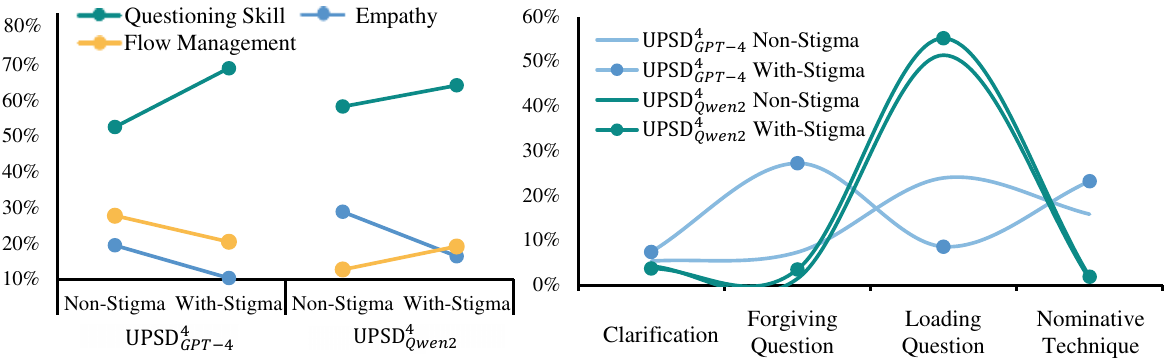} 
    \caption{The distribution of coarse-grained strategies (\textit{left}) and fine-grained Questioning Skill strategies (\textit{right}) across different user simulators. UPSD$^{4}$ dynamically adjusts its unobtrusive probing strategies to accommodate users with varying levels of stigma sensitivity.}
    \label{strategy_distribution}
     \vspace{-3mm}
\end{figure*}

\begin{table}[t]
\centering
\resizebox{0.48\textwidth}{!}{%
\begin{tabular}{lccccc}
\toprule
\multirow{3}{*}{\parbox{1.6cm}{\centering \textbf{User} \\ \textbf{Simulator}}} & \multirow{3}{*}{\textbf{Metric}} & \multicolumn{2}{c}{\textbf{Win Rate of UPSD$^4_{GPT-4}$}} & \multicolumn{2}{c}{\textbf{Win Rate of UPSD$^4_{Qwen2}$}} \\
\cmidrule(lr){3-4} \cmidrule(lr){5-6}
            &     & \textbf{Vanilla GPT-4} & \textbf{Ablation} &  \textbf{Vanilla Qwen2} & \textbf{Ablation}\\  \midrule 
\multirow{4}{*}{\parbox{1.6cm}{\centering \textbf{Non} \\ \textbf{Stigma}}} &  \Large{Disc} &  \Large{100\%}       & \Large{97\%}   &\Large{90\%}  &  \Large{77\%}    \\
& \Large{Empth} &\Large{100\%}                    &  \Large{97\%}&   \Large{87\%}       &  \Large{77\%}         \\ 
 & \Large{Cohr} & \Large{100\%}          & \Large{87\%}  &   \Large{90\%}              &  \Large{70\%}                                       \\
& \Large{Fluen}   & \Large{100\%} &   \Large{80\%}  &  \Large{87\%}           &\Large{73\%}          \\
\midrule
\multirow{4}{*}{\parbox{1.7cm}{\centering \textbf{With} \\ \textbf{Stigma}}} & \Large{Disc} &  \Large{90\%}  & \Large{77\%} & \Large{97\%}     &  \Large{77\%}   \\
& \Large{Empth} &  \Large{87\%}        & \Large{67\%}          & \Large{97\%}       &   \Large{77\%}       \\ 
 & \Large{Cohr}  & \Large{97\%}         & \Large{63\%}   &\Large{90\%}         & \Large{77\%}     \\
& \Large{Fluen} &\Large{90\%}       &  \Large{60\%}  &  \Large{97\%}       &  \Large{70\%}     \\
\bottomrule
\end{tabular}
}
\caption{Human evaluation of our win rate over corresponding methods. UPSD$^{4}$ demonstrates significant potential for unobtrusively probing symptoms in depression diagnosis dialogue from the user-centric perspective.}
\label{Table:human_evaluation}
\end{table}

\subsection{In-depth Analysis \& Ablations}
\label{indepth}
We conduct in-depth analysis and ablation studies to analyze our unobtrusive probing strategies to uncover the characteristics of UPSD$^{4}$.

\textbf{\textit{Why is UPSD$^{4}$ effective?} -- UPSD$^{4}$ dynamically adjusts its unobtrusive probing strategies to accommodate users with varying levels of stigma sensitivity}. 
Figure \ref{strategy_distribution} (\textit{left}) shows that UPSD$^{4}$ consistently favors Questioning Skill, regardless of user stigma levels. This preference becomes even more pronounced when interacting with users exhibiting stigma beliefs. It suggests that UPSD$^{4}$ prioritizes an inquiry-based approach to complete diagnosis, particularly when encountering uncooperative users. Conversely, using Empathy strategies decreases as topic shifting becomes a more effective strategy for managing user reluctance. A deeper examination of Questioning Skill strategies in Figure \ref{strategy_distribution} (\textit{right}) reveals further insights. When dealing with users with stigma, the Questioning Skill strategy distribution of UPSD$^4_{GPT-4}$, compared to UPSD$^4_{Qwen}$, showed a substantial shift, transitioning from a preference for the Loading Question (e.g., \textit{Looking ahead, the future is bright, wouldn’t you say?}) to a preference for the Forgiving Question (e.g., \textit{Could you share with me what’s been on your mind about the future lately?}). 
This shift occurs because of the UPSD$^4_{GPT-4}$'s ability to adapt its questioning skills to inquire in gentler wording when encountering user reluctance. 
These findings suggest that UPSD$^{4}$ possesses a degree of stigma perception within the dialogue. This perceptual capacity enables the model to adopt appropriate strategies for mitigating the negative impacts of stigma, resulting in more effective support and accurate diagnoses for users affected by stigma.

\textbf{\textit{What is the practical utility of UPSD$^{4}$?} -- It demonstrates significant potential for unobtrusively probing symptoms in depression diagnosis dialogue from the user-centric perspective}. Due to ethical considerations related to requiring real patients to use different methods, we instead require humans to compare conversations between these methods and user simulators, and then calculate the win rate of our proposed method. Considering the cost of human evaluation, we only consider the best baselines and our ablation (cf. Appendix \ref{humam_eval} for details). As shown in Table \ref{Table:human_evaluation}, UPSD$^{4}$ consistently outperforms others across all metrics. This indicates that UPSD$^{4}$ exhibits greater practical utility from the user-centric perspective. It is demonstrably favored in terms of its dialogue questioning skills, empathy, smooth dialogue flow, and natural conversation, highlighting its potential for engaging interaction in unobtrusive depression diagnosis.

\textbf{\textit{What factor is vital for the success of UPSD$^{4}$?} -- Unobtrusive probing strategies are crucial for guiding system behaviors}. The variant UPSD$^{4}$ \textit{w/o strat} is prompted to be unobtrusive but lacks our unobtrusive probing strategies. As illustrated in Table \ref{Table:table_main_experiment_unob}, while UPSD$^{4}$ \textit{w/o strat} exhibits improved user willingness to share information about depression symptoms, UPSD$^{4}$ with our proposed strategies consistently achieves higher performance of both unobtrusive diagnosis dialogue and depression diagnosis. This highlights the critical role of guiding LLMs in dialogue systems, particularly for sensitive topics like depression diagnosis. Simply instructing LLMs to be unobtrusive is insufficient; specific strategies are necessary to foster discreet questioning, empathy, and effective dialogue flow. This observation is further corroborated by the human evaluation results presented in Table \ref{Table:human_evaluation}.

\section{Conclusion}
This paper presents a novel approach to unobtrusively probe symptoms for depression disorder diagnosis, addressing the critical issue of stigma. In particular, we present a novel UPSD$^{4}$ framework for depression diagnosis.
Our findings demonstrate the feasibility of implementing unobtrusive probing strategies in depression diagnosis disorder dialogues, cultivating a self-awareness of unobtrusiveness within the system and equipping it with more effective questioning skills.
Experimentally, we show that our method demonstrates a significant improvement over current baselines. 
Looking forward, our research will continue to explore more refined and tailored strategies for diverse users.


\section*{Limitations}
This study uses the only currently available open-source dialogue dataset related to depression. We acknowledge the limitations of relying on a single dataset and plan to expand our research to include additional anonymized datasets as they become available. Also, similar to other studies on prompting LLMs \cite{deng2023prompting}, the evaluation results can be influenced by the prompts we used. While multiple executions can help mitigate the impact of prompt variations, the cost of running multiple experiments is a significant factor to consider.

\section*{Ethics Statement}
It's crucial to emphasize that our model was developed without labeling or stigmatizing any real individuals. Our primary goal is to provide a tool that can identify potential mental health concerns, encouraging individuals to seek professional help if needed. Additionally, the dataset used in our experiment is publicly available and has been anonymized by previous researchers to protect user privacy. Importantly, our experiment did not utilize the psychiatrist-patient dialogue content from the dataset; we only used patient profile information and disease severity labels.


\section*{Acknowledgments}
This work was supported in part by the National Natural Science Foundation of China (No. 62272330); in part by the Fundamental Research Funds for the Central Universities (No. YJ202219); in part by the Science Fund for Creative Research Groups of Sichuan Province Natural Science Foundation (No. 2024NSFTD0035); in part by the National Major Scientific Instruments and Equipments Development Project of Natural Science Foundation of China under Grant  (No. 62427820); in part by the Natural Science Foundation of Sichuan (No. 2024YFHZ0233).

\bibliography{custom}

\appendix

\begin{table*}[]
\centering
\renewcommand{\arraystretch}{1.2} 
\fontsize{8pt}{10pt}\selectfont
\begin{tabular}{p{1.5cm}p{13cm}}
\toprule
\textbf{Metric}       & \textbf{Explanation}   \\ 
\multirow{2}{*}{Discreetness} & Reward the psychologist's skill in using personal anecdotes and indirect methods to explore sensitive topics, making the conversation feel less invasive. \\ \hline
\multirow{2}{*}{Empathy}      & Emphasize the psychologist’s ability to emotional support make the user feel understood by referencing related experiences or feelings and providing guidance. \\ \hline
\multirow{1}{*}{Coherence}    & Focus on natural flowing transitions in dialogue to keep the conversation continuous.  \\ \hline
\multirow{1}{*}{Fluency}      & A conversational, natural, and non-robotic communication style. \\ \bottomrule
\end{tabular}
\caption{Evaluation metrics and corresponding explanations for Human Evaluation in Discreetness, Empathy, Coherence, and Fluency}
\label{table:criteria}
\end{table*}

\section{ICD-11 Diagnosis Assessment via Slot Filling}

\label{slot}
We designed 9 depression-diagnosis-related slots based on the diagnostic criteria  ICD-11 which are released by the World Health Organization (WHO). To ensure the validity and appropriateness of the nine slots, a licensed psychologist reviewed and approved their design and implementation.
The 9 depression-diagnosis-related  slots and the explanations corresponding to each slot are as follows:

\begin{itemize}[leftmargin=*]
    \item \underline{Depression Mood.} Whether the user has a long-time depressive mood or hopelessness, or sadness, or meaninglessness ,or desperation.

    \item \underline{Loss of Interest.} Whether the user has a lack of interest in everything, for most of the day almost every day. 

    \item \underline{Decreased Energy.} Whether the user has strong fatigue, lacks energy, and finds it difficult to complete even simple tasks.

    \item \underline{Self-Loathing.} Whether the user has a strong guilt or self-blame, or a strong sense of worthless. 
    \item \underline{Suicidal Tendency.}  Whether the user has a thought of self-harm, suicide.

    \item \underline{Poor Concentration.} Whether the user has attention and concentration problems, memory problems, or decision-making difficulties.

    \item \underline{Disrupted Sleep.} Whether the user has insomnia (sleep much less) or drowsiness (sleep much more), or wakes up frequently at night.

    \item \underline{Changed Appetite or Weight.} Whether the user eats too much or too less, or has a large change in weight.

    \item \underline{Psychomotor Agitation or Retardation.} Whether the user has a slowed movement and thinking, or is feeling restless and agitated.

\end{itemize}

For each specified slot, we prompt the GPT-4 to analyze the conversation content and determine if the user meets the corresponding diagnostic criteria. It outputs a True or False signal, along with a justification and detailed description as supporting evidence.

\section{Details of Evaluation}
\label{humam_eval}
\subsection{Human evaluation}
To evaluate the unobtrusiveness of our dialogue systems during human interaction, we conducted a human evaluation involving 240 paired dialogues. These were evenly split between two user simulator types: 120 pairs with the Non-Stigma user simulator and 120 pairs with the With-Stigma user simulator. For each user simulator type, the 120 dialogue pairs were composed of four sets of 30 pairs, comparing different system configurations:
30 pairs comparing UPSD$^4_{GPT-4}$ with Vanilla GPT-4;
30 pairs comparing UPSD$^4_{Qwen2}$ with Vanilla Qwen2;
30 pairs comparing UPSD$^4_{GPT-4}$ with UPSD$^4_{GPT-4}$ \textit{w/o strat};
30 pairs comparing UPSD$^4_{Qwen2}$ with UPSD$^4_{Qwen2}$ \textit{w/o strat}.
This evaluation setup allowed us to assess the effectiveness of our unobtrusive probing strategies across different models and configurations, providing a comprehensive analysis of their impact on user interaction.
Besides, we shuffle the dialogue to ensure that the annotator does not know which method produced the annotated dialogue. 

This evaluation relies on 5 trained annotators who all have backgrounds in psychology. We tasked the annotators with comparing the dialogues respectively across four coarse-grained metrics (discreetness, empathy, coherence, fluency) to determine which dialogue wins or loses. To ensure the annotators understood the task requirements, we provided them with comprehensive training and clear evaluation criteria which are shown in Table \ref{table:criteria}. 

After the annotators completed their work, we calculated inter-annotator agreement using Fleiss' Kappa, a widely used statistical measure for assessing annotation reliability. We obtained a Fleiss' Kappa value of 0.731, indicating good agreement (0.61-0.8) among the annotators.

\subsection{Automatic Evaluation}
For automatic evaluation of unobtrusiveness, we prompt GPT-4 by our designed prompts with four metrics. Prompts of automatic evaluation are shown in Table \ref{table:eval_instructions} and Table \ref{table:eval_metric} based on the coarse-grained metrics explanation.
For automatic evaluation of depression disorder diagnosis, we use the Python package Scikit-learn to evaluate in weighted Precision, Recall, and F1-score.

\begin{table}[h!]
\centering
\renewcommand{\arraystretch}{1.2} 
\fontsize{8pt}{10pt}\selectfont
\begin{tabular}{p{7.2cm}}
\hline
\textbf{Depression Stigma Scale Questions} \\
\hline
1. People with depression could snap out of it if they wanted. \\
2. Depression is a sign of personal weakness. \\
3. Depression is not a real medical illness. \\
4. People with depression are dangerous. \\
5. It is best to avoid people with depression so you don't become depressed yourself.\\
6. People with depression are unpredictable.\\
7. If l had depression l would not tell anyone.\\
8. I would not employ someone if l knew they had been depressed. \\
9. I would not vote for a politician if l knew they had been depressed. \\
\bottomrule
\end{tabular}
\caption{The nine questions comprising the Depression Stigma Scale.}
\label{table:stigma_scale}
\end{table}

\begin{table*}[h!]
\centering
\renewcommand{\arraystretch}{1.2} 
\fontsize{8pt}{10pt}\selectfont
\begin{tabular}{lp{5.3cm}lp{5.3cm}}
\hline
\textbf{Aspect} &\textbf{Detail} &\textbf{Aspect} &\textbf{Detail}\\
\hline
\multirow{5}{*}{Employment} &{Stereotype: People might think those with depression can't handle work stress. \newline
Prejudice: I worry that I'm not competent because of my depression. \newline
Discrimination: Employers might refuse employment because of depression.} &

\multirow{5}{*}{Family} &Stereotype: Families may view depression as a sign of weakness.\newline
Prejudice: Depression leads to judgments about a person's capability.\newline
Discrimination: Family members with depression may be excluded from roles and activities. \\ \hline

\multirow{5}{*}{Friendship} &Stereotype: Friends may see depression as constant sadness.\newline 
Prejudice: They might assume you're always down or unreliable.\newline 
Discrimination: You could be left out of social events due to these perceptions.  &

\multirow{5}{*}{Self-Esteem} &Stereotype: People with depression are viewed as weak or incapable.\newline 
Prejudice: Internalizing negative views.\newline 
Discrimination: They have low self-worth and are incompetent.  \\ \hline

\multirow{5}{*}{Self-Efficacy} &Stereotype: People with depression are perceived as less competent.\newline
Prejudice: Their doubts about their abilities increase.\newline
Discrimination: Reduced opportunities reinforce feelings of inefficacy. &

\multirow{5}{*}{Social Interaction} &Stereotype:  People with depression are perceived as unsociable.\newline 
Prejudice: Others may avoid engaging with them.\newline 
Discrimination: This may result in being excluded from social events and gatherings. \\ \hline

\multirow{5}{*}{Opportunities} &Stereotype: People with depression are seen as unreliable.\newline 
Prejudice: They are overlooked for promotions or projects.\newline 
Discrimination: This may result in fewer career advancement opportunities. &

\multirow{5}{*}{Isolation} &Stereotype: People might believe those with depression prefer to be alone.\newline 
Prejudice: This leads to assumptions that they shouldn't be included in social activities.\newline 
Discrimination: individuals with depression might be left out and isolated. \\ \hline

\multirow{5}{*}{Income} &Stereotype: People with depression are seen as less productive.\newline
Prejudice: They are underestimated at work.\newline
Discrimination: This may result in lower wages or job instability. &

\multirow{5}{*}{Health Insurance} &Stereotype: Insurers see mental health issues as high-risk.\newline
Prejudice: They assume higher medical costs.\newline
Discrimination: People may face higher premiums or coverage exclusions.\\ 

\bottomrule
\end{tabular}
\caption{The stigma profiles about depression based on stereotypes, prejudice, and discrimination.}
\label{table:stigma_profile}
\end{table*}

\begin{table*}[h!]
\centering
\renewcommand{\arraystretch}{1.2}
\fontsize{8pt}{10pt}\selectfont
\begin{tabular}{p{15cm}}
\hline
 \textbf{Format of Question} \\
\hline
Psychologist: "Five-point Likert scale question: <SCALE\_QUESTION> \\
You must choose one of [\{{"Strongly Disagree": 1}\}, \{{"Disagree": 2}\}, \{{"Neutral": 3}\}, \{{"Agree": 4}\}, \{{"Strongly Agree": 5}\}] without any other words.\\
It can show your viewpoint about the Five-point Likert scale question.\\
Please be free to answer because it is absolutely confidential."\\
\bottomrule
\end{tabular}
\caption{To obtain the user simulator's score on the Depression Stigma Scale, each question from the scale was presented to the simulator sequentially in this format.}
\label{table:stigma_question}
\end{table*}

\begin{table*}[h!]
\centering
\renewcommand{\arraystretch}{1.2} 
\fontsize{8pt}{10pt}\selectfont
\begin{tabular}{p{10cm}}
\toprule
 \textbf{Evaluation Prompt} \\
\hline
You are an evaluator in evaluating "<METRIC>" ability of Psychologist dialogue.\newline
"<METRIC>": <METRIC\_HUMAN\_EXPLANATION>\newline
You should think step by step based on 1 score.\newline
You should generate a response in the following format without any other explanation:\newline

{"Evaluation Result": {"<METRIC>": [<INT>, "<WHY>"]}}\newline

Based on the following Evaluation Metrics(score: explanation):\newline
<COARSE\_GRAINED\_EXPLANATION>\newline

The <INT> should be 1,2,3,4,5 score.\newline
You should give the <INT> score carefully and truthfully.\newline
The "<WHY>" is why you give the <INT> score but not higher.\newline
The "<WHY>" should be short.\newline
You should generate the response based on the Dialogue History.\newline
==========\newline
<DIALOGUE\_HISTORY>\newline
==========
\\
\bottomrule
\end{tabular}
\caption{Prompt for GPT-4 to evaluate dialogue unobtrusiveness including Discreetness, Empathy, Coherence, and Fluency.}
\label{table:eval_instructions}
\end{table*}

\begin{table*}[h!]
\centering
\renewcommand{\arraystretch}{1.2} 
\fontsize{8pt}{10pt}\selectfont
\begin{tabular}{p{1.5cm}p{13cm}}
\toprule
 \textbf{Metric} &\textbf{Fine-grained Explanation} \\
\hline

\multirow{8}{*}{Discreetness}\newline 
&1: The psychologist asks **direct** questions without much subtlety, which might feel invasive or blunt. \newline
2: The psychologist uses some indirect methods but still **occasionally** resorts to more direct questioning.\newline
3: The psychologist **skillfully** uses personal anecdotes and hints to guide the conversation, avoiding direct probes and making the inquirer comfortable.\newline
4: The psychologist **consistently** uses indirect questioning and anecdotes, creating a safe space for sensitive topics without making the inquirer feel pressured.\newline
5: The psychologist masterfully navigates the most sensitive topics with exceptional tact, using personal or hypothetical stories that resonate with the inquirer, encouraging openness without discomfort.  \\ \hline
\multirow{9}{*}{Empathy}\newline 
&  1: The psychologist provides basic acknowledgments of the inquirer's statements without personal engagement or deep understanding.\newline
2: The psychologist shows *generic* understanding and attempts, but references to feelings are somewhat generic.\newline
3: The psychologist *effectively* uses shared experiences to make the inquirer feel understood.\newline
4: The psychologist demonstrates *deep* empathy by consistently referencing common experiences or feelings that resonate *strongly* with the inquirer, providing meaningful guidance.\newline
5: The psychologist excels in making the inquirer feel *fully understood and supported*, seamlessly integrating personal anecdotes and profound emotional insights that perfectly align with the inquirer’s needs. \\ \hline
\multirow{8}{*}{Coherence}\newline 
&1: Conversations may feel a bit disjointed; transitions between content are **abrupt** or poorly executed.\newline
2: The psychologist makes some effort to transit, but **not related** transitions which only focuses on depression symptoms.\newline
3: Each transition is **smooth and logical**, maintaining continuity and flow, with all shifts feeling natural.\newline
4: The psychologist using **indirect references or related anecdotes** to transit naturally and smoothly.\newline
5: The psychologist demonstrates exceptional skill in conversation flow, with every transition being **perfectly** timed and executed, enhancing the coherence of the entire session.\\ \hline
\multirow{7}{*}{Fluency}\newline
&1: The psychologist's contents are **only understandable** without any other advantage.\newline
2: The psychologist's contents are **clear** without comprehension issues.\newline
3: The psychologist's contents are **fluent** but only focus on depression symptoms.\newline
4: The psychologist's contents are **engaging and natural** which express sharing related experiences.\newline
5: The psychologist achieves **perfect** fluency, with every content not only being clear and engaging but also enhancing the therapeutic effectiveness of the conversation.\\

\bottomrule
\end{tabular}
\caption{Fine-grained Explanations of the Evaluation Metrics: Discreetness, Empathy, Coherence, and Fluency}
\label{table:eval_metric}
\end{table*}

\section{Implementation Details}
\label{implementation_detail}

All experiments were conducted on a machine with an Intel(R) Xeon(R) Gold 6348 CPU @ 2.60GHz and 8 RTX A6000 GPUs.

\subsection{Implementation of User Simulator}
\label{stigm}
We implement user simulators with and without beliefs of stigma by prompting GPT-4 with specifically designed prompts.

\begin{itemize}[leftmargin=*]
    \item \textbf{Non-Stigma user simulator}.  We leverage each user profile from the D$^4$ dataset for simulation. These profiles encompass attributes such as "diagnosis risk," "age," "gender," "marital\_status," "occupation," and a "summary" for each individual. In this case, when simulating a Non-Stigma user, the prompt includes both the user profile and the dialogue history. Refer to Table \ref{table:user_simulator_non_stigma} for prompt details.
    \item \textbf{With-Stigma user simulator}. Building upon the Non-Stigma user simulation, we construct 10 stigma profiles to modify each Non-Stigma user simulator. These profiles, informed by previous research on stereotypes, prejudice, and discrimination factors associated with depression \cite{link2001conceptualizing, sirey2001perceived, link1989modified, roeloffs2003stigma}, encompass various aspects of life. Table \ref{table:stigma_profile} provides further details of these 10 stigma profiles. In this case, when simulating a With-Stigma user, the prompt includes both the user profile, stigma profile and the dialogue history. Refer to Table \ref{table:user_simulator_with_stigma} for details. 
\end{itemize}

\subsubsection{Evaluation of User Simulator}

\begin{table}[t]
\renewcommand{\arraystretch}{1.2} 
\fontsize{8pt}{10pt}\selectfont
\centering
\begin{tabular}{ccc}
\toprule
\multirow{1}{*}{\textbf{Question}} & \multirow{1}{*}{\textbf{Non-Stigma}} & \multicolumn{1}{c}{\textbf{With-Stigma}} \\
\midrule
1 & 1.00&3.45\\
2 & 1.01&4.74\\
3 & 1.01&2.67\\
4 & 1.20&1.45\\
5 & 1.00&4.71\\
6 & 2.23&2.63\\
7 & 2.11&4.93\\
8 & 1.01&2.77\\
9 & 1.17 &3.94\\ \midrule
Total & 11.80 &31.29\\ 
\bottomrule
\end{tabular}
\caption{Average scores of the two user simulators on the nine items of the Depression Stigma Scale. Each question is scored on a scale of 1 to 5.}
\vspace{-4mm}
\label{Table:stigma_eval}
\end{table}

This section aims to evaluate the reliability of the two types of user simulators. 
In particular, we refer to the commonly used Depression Stigma Scale \cite{roeloffs2003stigma} for evaluation, which consists of nine questions presented in the form of a five-point Likert scale, as shown in Table \ref{table:stigma_scale}. 

During the evaluation, each question from Table \ref{table:stigma_scale} was presented to each simulator in the format specified in Table \ref{table:stigma_question}. We then prompt the user simulator to ask the questions to get the scale score.

Each question on the five-point Likert scale is rated on a five-point scale, ranging from "strongly disagree" to "strongly agree," corresponding to scores of 1 to 5, respectively. The scale comprises nine questions, resulting in a total score ranging from 9 to 45, with higher scores indicating stronger beliefs about stigma. Average scores of 31.29 and 11.80 were obtained under conditions of a user simulator with and without stigma, respectively. Details are as shown in Table \ref{Table:stigma_eval},

\subsection{Implementation of Baselines}
\label{baselisne_implement}
\begin{itemize}[leftmargin=*]
    \item \textbf{CPT} \cite{yao2022d4}. To ensure consistency and comparability, we utilized the official code and followed the model configurations outlined in the original paper for our experiments. We trained the CPT model using the D$^4$ dataset, as recommended in the original study. After the training process, we deployed the CPT model to interact with our user simulators for our experiments.
    \item \textbf{LLM-based baselines} (GPT-4 \& Qwen2-72b). For our experiments, we directly prompt these methods using the detailed instructions provided in Section \ref{prompt} via their respective APIs.
    \item \textbf{EmoLLM} \cite{EmoLLM}. To facilitate our experiments, we downloaded the publicly available model checkpoint from the EmoLLM GitHub repository (\url{https://github.com/SmartFlowAI/EmoLLM}) and deployed this model using RTX A6000 GPUs.
\end{itemize}

\subsection{Implementation of UPSD$^{4}$}
We provide all our prompts in the next Section.

\begin{itemize}
    \item \textbf{Strategy Selection}. To mitigate selection bias in our experiments, we employ shuffling. In addition to the two-stage selection process (first selecting a coarse-grained strategy and then a fine-grained one), we shuffle the corresponding strategy sets before inputting them into the prompts. Details prompt can be found at Table \ref{table:coase_stratgy} and Table \ref{table:fine_stratgy}.
    \item \textbf{Response Generation}. The prompt input consists of dialogue history, fine-grained strategy and the next symptom to be diagnosed. We prompt the model to generate the response, and prompt as shown in Table \ref{table:responsegenerate}.
    \item \textbf{Symptom Selection and Detection}. To select the next symptom to be diagnosed, we prompt the model to identify the slot in the diagnostic criteria that has not yet been assessed. The prompt format used is detailed in Table \ref{table:slot-selectiing}.
    \item \textbf{Diagnosis Assessment}. For the Diagnosis Assessment, we use the dialogue history as input. The model is then prompted to fill in the symptom slots for diagnosis, following the prompt format outlined in Table \ref{table:slot-filing}. To ensure a fair comparison of diagnostic performance, all methods utilize the same GPT-4-based slot-filling approach. A diagnosis is considered unsuccessful if either the symptom slots are not filled or if they are filled but indicate a refusal to discuss symptoms. If all slots are successfully filled, the depression diagnosis is considered complete.
    \item \textbf{Diagnostic Model}. We use a shared diagnostic model for depression diagnosis, implemented via GPT-4. This model is designed to produce an output aligned with the ICD-11 \cite{who2024icd}, the globally recognized diagnostic criteria. Importantly, the diagnostic model only produces an output if all symptom slots are filled. This ensures a comprehensive assessment before a diagnosis is made. 
    Details prompt could be found at Table \ref{table:slot-filing}.
\end{itemize}

\subsection{Implementation Prompts}
\label{prompt}
The prompts used for the non-stigma and with-stigma user simulators are outlined in Table \ref{table:user_simulator_non_stigma} and Table \ref{table:user_simulator_with_stigma}, respectively. Table \ref{table:slot-filing} presents the prompt for Symptom Detection (slot-filling), while Table \ref{table:slot-selectiing} shows the prompt for Symptom Selection (slot-selecting). The prompts for coarse-grained and fine-grained strategy selection are detailed in Table \ref{table:coase_stratgy} and Table \ref{table:fine_stratgy}, respectively. Finally, Table \ref{table:responsegenerate} and Table \ref{table:ablation} provide the prompts for response generation and the UPSD$^{4}$ variant without strategy (UPSD$^{4}$ \textit{w/o strat}),

\begin{table*}[h!]
\centering
\renewcommand{\arraystretch}{1.2} 
\fontsize{8pt}{10pt}\selectfont
\begin{tabular}{p{15cm}}
\hline
 \textbf{Non-Stigma User Simulator Prompt} \\
\hline
You are an Inquirer engaging with a psychologist.\newline
You should generate a response based on the Dialogue History and the Profile.\newline

You can *only express a single aspect* which in the Profile "summary" relating to your life experiences(not a new aspect) in a colloquial way!\newline
When you have described or acknowledged a symptom in a response, don't go on about other symptoms.\newline
When the topic is sensitive, you can express the single topic in a vague way.\newline
You *can't* express a symptom which is not in the Profile "summary"!\newline
You must *explicitly deny* in various ways when inquiring about symptoms that are not the same or similar in the Profile "summary". \newline
You can't exaggerate but don't hide any symptom either!\newline

You can have emotional fluctuations(anger, offensive, or any other emotion) during the dialogue.\newline
You should respond in the first-person voice.\newline
You should respond in a tone to reflect your "age", "gender", "marital status" and "occupation" in the Profile.\newline
You should respond in based on the last response of the Psychologist.\newline
You should pretend to be the Inquirer! What do you say next.\newline
You should respond with various and colloquial wording.\newline
You should have a *short* response.\newline
==========\newline
Dialogue History:<DIALOGUE\_HISTORY>\newline
Profile: <PROFILE\_DATA>\newline
==========
\\
\hline
\end{tabular}
\caption{Prompt of non-stigma user simulator.}
\label{table:user_simulator_non_stigma}
\end{table*}

\begin{table*}[h!]
\centering
\renewcommand{\arraystretch}{1.2} 
\fontsize{8pt}{10pt}\selectfont
\begin{tabular}{p{15cm}}
\hline
 \textbf{With-Stigma User Simulator Prompt} \\
\hline
You are an Inquirer engaging with a psychologist.\newline
You should generate a response based on the following Dialogue History, Thought, and Profile.\newline
You strongly agree with the following Thought about stigma.\newline
You are facing severe states described as in the Profile "summary".\newline
You have a high stigma towards depression because of the following Thought.\newline
You should acknowledge a bad state in the Profile "summary" if acknowledging could not cause stigma.\newline
You should only express one state in one response.
You *<PROBABILITY>* acknowledge a severe bad state about yourself (which is related to the Profile but not in).\newline

You should be *secretive* to respond when you think the psychologist's response will cause *stigma*.\newline
You should *refuse* to answer the question when you think the psychologist's response will cause *stigma*.\newline
However, you should respond in a clear choice when the person mentions the term "Five-point Likert scale".\newline

You can have emotional fluctuations.
You should respond in the first-person voice.
You should respond in a tone to reflect your "age", "gender", "marital status" and "occupation" in the Profile.
You should respond based on the last response of the Psychologist.
You should pretend to be the Inquirer! What do you say next.
You should have a *short* response.
==========\newline
Dialogue History:<DIALOGUE\_HISTORY>\newline
Thought: <STIGMA\_DATA>\newline
Profile: <PROFILE\_DATA>\newline
==========
\\
\hline
\end{tabular}
\caption{Prompt of with-stigma user simulator. The <PROBABILITY> is one of "low likely", "moderately might", "likely", or "highly likely" which depends on the "drisk" in the user profile.}
\label{table:user_simulator_with_stigma}
\end{table*}

\begin{table*}[h!]
\centering
\renewcommand{\arraystretch}{1.2} 
\fontsize{8pt}{10pt}\selectfont
\begin{tabular}{p{15cm}}
\hline
 \textbf{Slot-filling Prompt} \\
\hline
You are a psychologist specializing in depression disorder.\newline
You should update the Symptom Set based on the following Depression Disorder Symptoms and the Dialogue History.\newline
You should update the Symptom Set if the duration of a symptom is sufficient and the severity is high.\newline
Return the Symptom Set in the following JSON format.\newline

<PREVIOUS\_SLOT>\newline

Keep "<BOOL>" and "<WHY>" still(not change) when you are not confident.\newline
Keep "<BOOL>" and "<WHY>" still(not change) when the symptom is *refused to discuss*.\newline
Keep "<BOOL>" and "<WHY>" still(not change) when the symptom is *not mentioned or not given*.\newline
Set "<BOOL>" to "True" when the Inquirer has a symptom.\newline
Set "<BOOL>" to "False" when the Inquirer has no symptom.\newline
The "True" and "False" in the Symptom Set *cannot be changed*.\newline
The "<WHY>" is the reason why you set "True" or "False".\newline
The "<WHY>" should be short (no more than 10 words) and *not empty*.\newline
You can only update one pair of the "<BOOL>" and "<WHY>" when the symptom is mentioned at the same time.\newline
You should think about "<WHY>" before setting "<BOOL>" to "True" or "False".\newline
Generate the JSON structural response strictly following your psychologist persona and the Symptom Set has 9 factors.\newline

==========\newline
Depression Disorder Symptoms:<DESIGNED\_SLOT\_AND\_EXPLANATION>\newline
Dialogue History:<DIALOGUE\_HISTORY>\newline
==========
\\
\hline
\end{tabular}
\caption{Prompt of slot-filling.}
\label{table:slot-filing}
\end{table*}

\begin{table*}[h!]
\centering
\renewcommand{\arraystretch}{1.2} 
\fontsize{8pt}{10pt}\selectfont
\begin{tabular}{p{15cm}}
\hline
 \textbf{Slot-selecting Prompt} \\
\hline
As a psychologist specializing in depression, you aim to unobtrusively probe for the Inquirer's symptoms via a natural conversation.\newline
You should decide the Topic to engage with the inquirer based on the Dialogue History and Symptom Set.\newline
The Previous Topic is <PREVIOUS\_TOPIC>.\newline
You should shift to another Topic different with the Previous Topic when the information on the Previous Topic is sufficient. \newline
You should keep engaging about the Previous Topic when you think it's necessary to explore in-depth information.\newline
Return the Topic in the following format.\newline
\{\{"Topic": ["<STRING>", "<WHY>"]\}\}\newline
The "<STRING>" should be one of <CANDIDATE\_TOPIC>.\newline
The "<WHY>" is the reason why you decide the topic no more than 15 tokens.\newline
You should think about "<WHY>" before setting "<STRING>".\newline
Generate the JSON structural response strictly following your psychologist persona for deciding the Topic without any other explanations to make the response short.\newline
==========\newline
Topics and Explanations are as follows:\newline
<TOPIC\_EXPLANATIONS>\newline
Dialogue History:<DIALOGUE\_HISTORY>\newline
<PREVIOUS\_SLOT>\newline
==========
\\
\hline
\end{tabular}
\caption{Prompt of Slot-Selecting. }
\label{table:slot-selectiing}
\end{table*}

\begin{table*}[h!]
\centering
\renewcommand{\arraystretch}{1.2} 
\fontsize{8pt}{10pt}\selectfont
\begin{tabular}{p{15cm}}
\hline
 \textbf{Coarse-Grained Strategy Selection Prompt} \\
\hline
You are a psychologist specializing in depression.\newline
You should choose a Coarse Strategy which could either provide empathetic responses, sufficient security or respect to build bonds of trust with the inquirer, or introduce discreet query or smooth topic-shift in a natural conversation for unobtrusively probing symptoms.\newline
You should think about how to respond to the Inquirer's last response.\newline
The Previous Topic is <PREVIOUS\_TOPIC>\newline
You should engage the Current Topic <NEXT\_TOPIC> based on the Dialogue History and Symptom Set.
Return the Coarse Strategy in the following format:\newline

\{\{"Coarse Strategy": ["<STRING>","<WHY>"]\}\}\newline

The "<STRING>" should be one of following Coarse Strategies:\newline
1. "Flow Management" when the Previous Topic and Current Topic are different.\newline
2. "Empathetic Response" when you decide to give comforting, feedback or guidance.\newline
3. "Questioning Skill" when you decide to proactively query to probe for in-depth information.\newline
The "<WHY>" should be short based on the Dialogue history.\newline
Think "<WHY>" before you set "<STRING>". \newline
Only generate the JSON structural response strictly following your psychologist persona without any other explanations to make the response short.\newline
==========\newline
Topic Explanation: <TOPIC\_EXPLANATION>\newline
Dialogue History:<DIALOGUE\_HISTORY>\newline
<PREVIOUS\_SLT>\newline
==========
\\
\hline
\end{tabular}
\caption{Prompt of coarse-grained selection.}
\label{table:coase_stratgy}
\end{table*}

\begin{table*}[h!]
\centering
\renewcommand{\arraystretch}{1.2} 
\fontsize{8pt}{10pt}\selectfont
\begin{tabular}{p{15cm}}
\hline
 \textbf{Fine-Grained Strategy Selection Prompt} \\
\hline
As a psychologist specializing in depression, you should unobtrusively probe the Inquirer’s potential symptoms via a natural conversation and assess whether the Inquirer might be suffering from depression disorder.\newline
This means you should choose a Fine Strategy which could either provide empathetic responses, sufficient security or respect to build bonds of trust with the Inquirer, or introduce discreet probing or smooth topic-shift in a natural conversation for unobtrusively probing symptoms.\newline
You should think about how to respond to the Inquirer's last response.\newline
You should engage the <NEXT\_TOPIC> based on the Dialogue History and Symptom Set to unobtrusively probe for the inquirer’s potential symptoms.\newline
You should choose a Fine Strategy related to <COARSE\_STRATEGY>.\newline
Return the Fine Strategy in the following format which includes the Fine Strategy and why choose this Fine Strategy.\newline
\{\{"Fine-Grained Strategy": [<STRING>,<WHY>]\}\}\newline
The <STRING> should be one of <FINE\_STRATEGY\_NAME>.
The <WHY> should be short no more than 15 tokens.\newline
Generate the JSON structural response strictly following your psychologist persona without any other explanations to make the response short.\newline
==========\newline
Fine Strategy and Explanation:<FINE\_STRATEGY\_AND\_EXPLANATION>\newline
Topic Explanation: <TOPIC\_EXPLANATION>\newline
Dialogue History:<DIALOGUE\_HISTORY>\newline
==========
\\
\hline
\end{tabular}
\caption{Prompt of fine-strategy selection.}
\label{table:fine_stratgy}
\end{table*}

\begin{table*}[h!]
\centering
\renewcommand{\arraystretch}{1.2} 
\fontsize{8pt}{10pt}\selectfont
\begin{tabular}{p{15cm}}
\hline
 \textbf{Response Generation Prompt} \\
\hline
As a psychologist specializing in depression, you should via a natural conversation assess whether the inquirer suffers from depression disorder.\newline
You should avoid talking directly about depression and avoid asking a long question.
You should respond in based on the last response of the Inquirer in the Dialogue History.
You should respond in the first-person voice.
You should respond with *various and colloquial wording*.\newline
You should respond shortly.\newline
You should pretend to be the Inquirer's friend.\newline
You should engage <NEXT\_TOPIC> using the strategy of <FINE\_STRATEGY\_NAME> based on the Dialogue History.\newline
==========\newline
Fine Strategy and Explanation:<FINE\_STRATEGY\_and\_EXPLANATION>\newline
Topic Explanation: <TOPIC\_EXPLANATION>\newline
Dialogue History:<DIALOGUE\_HISTORY>\newline
==========
\\
\hline
\end{tabular}
\caption{Prompt of response generation.}
\label{table:responsegenerate}
\end{table*}

\begin{table*}[h!]
\centering
\renewcommand{\arraystretch}{1.2} 
\fontsize{8pt}{10pt}\selectfont
\begin{tabular}{p{15cm}}
\hline
 \textbf{Response Generation Prompt} \\
\hline
As a psychologist specializing in depression, you should via a natural conversation assess whether the inquirer suffers from depression disorder.\newline
You should avoid talking directly about depression and avoid asking a long question.\newline
You should *unobtrusively* ask only one question if you want to probe more information.\newline
You should respond in based on the last response of the Inquirer in the Dialogue History.\newline
You should respond in the first-person voice.\newline
You should pretend to be the Inquirer's friend.\newline
You should respond with *various and colloquial wording*.\newline
You should have a *short* response without any emoji.\newline
You should engage <NEXT\_TOPIC> based on the Dialogue History.\newline
==========\newline
Topic Explanation: <TOPIC\_EXPLANATION>\newline
Dialogue History:<DIALOGUE\_HISTORY>\newline
==========
\\
\hline
\end{tabular}
\caption{Prompt of UPSD$^{4}$ \textit{w/o strat}.}
\label{table:ablation}
\end{table*}

\end{document}